# Measuring Digital Labour Market Transitions with a Digital Semantic Score: An AI-Based Methodology Applied to the Dutch Labour Market

Sadegh Shahmohammadi[1], Xavier Pinho[1], Mairi Bowdler[1], Suhendan Adiguzel-van Zoelen[1], Joost van Genabeek[1]

[1] Netherlands Organization for Applied Scientific Research TNO, P.O. Box 718, 2130 AS Hoofddorp, The Netherlands

## Abstract

The digital transformation of the Dutch labour market is reshaping occupational language, career pathways, and job-related skills. Addressing these changes requires granular labour market intelligence. This paper develops an AI-based methodology to analyse digitalisation using data covering millions of Dutch job profiles.

The methodology combines embedding-based similarity search and large language model classification to map unstructured job information to harmonised ESCO occupations. We also introduce a Digital Semantic Score that measures how strongly job titles and skills are associated with digital concepts relative to a non-digital reference. Using embeddings and cosine similarity to transparent digital and non-digital anchor groups, this indicator moves beyond keyword-based approaches by capturing broader digital meanings in occupational language and worker skill profiles. It enables analysis across occupations, career transitions, emerging job-title vocabulary, and skill digitality.

The findings reveal that digitalisation is unevenly distributed across the labour market. Digital job-title language is most prominent among managerial, professional and ICT-related occupations, but is increasingly visible in hybrid business, marketing and automation-related roles. Career-transition analyses show that movement toward digital work is pathway-dependent, while skill analyses highlight the multidimensional nature of digital capability, encompassing technical, hybrid and business-systems skills.

By combining profile data, AI-supported occupational classification and semantic scoring, this study advances AI-driven labour market analytics and provides a scalable framework for monitoring digital labour market change. The methodology helps identify emerging skill needs, support reskilling strategies, and inform policies addressing skills mismatches and labour shortages in the Netherlands.

## Introduction

The Dutch labour market is undergoing rapid transformation driven by digitalisation, automation and artificial intelligence (AI). In 2024, 22.7% of Dutch businesses with more than 10 employees integrated one or more types of AI technology (Statistics Netherlands [CBS]., 2025). These developments are changing the tasks performed within occupations, altering the demand for skills, and contributing to emerging labour and skills shortages (OECD., 2023; Green., 2024). Research on AI further suggests that its effects are unevenly distributed across occupational groups: Engberg et al. (2024) found that administrative and technical staff with relatively limited social interaction are disproportionately affected, while Webb (2020) highlights potential negative consequences for highly educated older professionals. At the same time, digital technologies are creating new occupational opportunities and changing the ways in which skills are combined and applied across jobs, strengthening the labour market position of those with the appropriate skills (Alekseeva et al., 2021). This includes those of the hybrid digital occupations, referred to as “laboratories” of skill rebundling (Stephany., 2021). This dynamic creates winners and losers in technological transformation, with workers in routine occupations often feeling disadvantaged (Frey & Osborne, 2013/2017).

While existing studies have generated important insights into occupational exposure to technology and the demand for digital skills (e.g. Green., 2024; Jensen & Schneider., 2026), important measurement challenges remain. Such studies often rely on broad occupational classifications, vacancy data or aggregate labour market statistics, which may not fully capture how digitalisation changes the language of work, the emergence of new job titles, or the transferability of skills across occupations. This is particularly relevant for studying labour and skill shortages, since shortages and mismatches are not only about the number of available workers, but also whether workers’ skills align with the changing occupational requirements.

Understanding these developments therefore requires data and methods capable of capturing labour market change at a more granular level and at a faster pace than many conventional data sources allow. Research is now increasingly utilizing the online labour market data, worker profiles and natural language processing techniques to analyse changing skill requirements, occupational dynamics and labour market mismatches at a level of detail that is difficult to achieve via conventional labour market statistics (Escudero et al., 2024; Mason et al., 2023). Such techniques have also been utilized to develop markers of occupational digitalization based on the digital skill requirements associated with different occupations (Lennon et al., 2023).

Recent advances in big-data labour market analytics and AI-supported text classification offer new opportunities to address these limitations. However, quickly evolving technologies such as AI continue to create challenges for identifying and classifying emerging digital skills using existing taxonomies and classification systems (Beblavý et al., 2016; Napierała, 2024).

Online profile data from professional network platforms now provides detailed information on job histories, occupational titles and reported skills, while embedding-based methods make it possible to measure semantic similarities between occupations, skills and digital concepts. Such methods can move beyond keyword-based approaches by capturing whether job titles and skills are semantically closer to digital work, hybrid digital work, business systems and automation, or more traditional non-digital work domains.

Building on these developments, this study uses large-scale online profile data for the Netherlands and AI-supported occupational classification to develop refined indicators of digital labour market change. The analysis links unstructured job-title and skill information to ESCO-based occupational classifications and introduces a new Digital Semantic Score to measure the degree to which job titles and skills are semantically associated with digital concepts. In doing so, the paper examines digitalisation not only as a characteristic of occupations, but also as a multidimensional labour market phenomenon reflected in occupational language, career mobility pathways and worker skill profiles.

The paper makes three main contributions. First, it develops an AI-supported approach for classifying large-scale online profile data into harmonised occupational categories suitable for labour market analysis. Second, it introduces a Digital Semantic Score that captures changes in the digital content of job titles and skills over time. Third, it applies these indicators to the Dutch labour market to examine how digitalisation varies across occupational groups, emerging job-title vocabulary, career transitions and high-prevalence skill profiles. Together, these analyses provide new evidence on where digitalisation is most visible, where it is spreading, and how targeted reskilling or mobility support may help address emerging skills mismatches.

# Data and Methods

## Dataset

The analysis uses large-scale online profile data provided by Revelio Labs, comprising over one billion job profiles, 20 million companies, more than 3,000 reported skills, job histories and educational information. Some fields of these data are directly extracted from LinkedIn (e.g. job title, skills), and other fields are predicted by Revelio Labs (e.g. gender, remote suitability).

The analysis presented in this study focuses on the Netherlands job-history data. The Netherlands ranks consistently as one of the highest markets globally for LinkedIn adoption (LinkedIn, n.d.) . The Netherlands also ranks top of the European Union's Digital Economy and Society Index (DESI), with over 83% of the population having basic to advanced digital skills, making the workforce's digital footprint unusually rich (European Commission, 2026). Each record of this dataset represents an occupational record for an individual LinkedIn profile. Each record contains a person identifier, a self-reported job title, start and end dates, an occupational code, and more.

For each unique user, the dataset contains a list of self-reported skills. While this enables analysis of individual users' skill profiles, the skills cannot be linked to specific jobs within a user's employment history. As a result, it is not possible to determine which skills were associated with particular occupations or to track how skills evolved across job transitions over time. This represents a limitation of the data, as it prevents individual longitudinal analyses of skill development and the examination of occupation-specific skill requirements throughout an individual's career trajectory.

## ESCO integration

A major challenge in the source data is that occupations are not directly coded in a way that is fully consistent with the European labour market context. Although the provider includes predicted O*NET (Occupational Information Network) codes, some of these assignments are not sufficiently accurate for the intended analyses. To improve comparability and support transparent and Europe-relevant indicators, occupations are therefore reclassified using the European Skills, Competences, Qualifications and Occupations framework (ESCO; (https://esco.ec.europa.eu), the European multilingual classification of occupations, skills and qualifications.

To map free-text job descriptions to ESCO occupations, we develop a two-stage AI-supported classification pipeline. In the first stage, vector embeddings are generated for both job descriptions and ESCO occupation entries. ESCO occupation representations are constructed from preferred labels, occupational descriptions, alternative labels, and hidden labels. Embeddings are generated using OpenAI's *text-embedding-3-large* model, producing 3,072-dimensional vector representations that capture the semantic content of occupational information and profile descriptions. Using the FAISS library for efficient nearest-neighbour search, ESCO occupation vectors are indexed and the ten most similar ESCO occupations are retrieved for each job profile based on vector similarity scores.

In the second stage, a large language model evaluates the original job description together with the top ten ESCO occupation candidates using a structured prompt and selects the most appropriate ESCO occupation code. When the model's confidence is insufficient, no occupational code is assigned. Prompt templates were iteratively refined to improve classification consistency across languages, sectors, and occupational domains. This hybrid approach combines the semantic breadth and high recall of embedding-based similarity search with the greater precision and contextual disambiguation capabilities of large language models. It is particularly suitable for multilingual and noisy text environments, where job titles may be ambiguous and profile descriptions sparse or inconsistent.

After assignment of ESCO occupations, occupational codes are harmonized to the four-digit ISCO level. ISCO stands for International Standard Classification of Occupations 2008 (ISCO-08). Non-numeric characters are removed from the original occupational code field and the first four digits are retained. The corresponding ISCO major group is defined as the first digit of the four-digit code. Records without a valid four-digit occupational code are excluded for all analyses requiring ISCO-based occupational classification.

## Career Transitions

Job-to-job transitions were constructed within individuals. For each person, all job records were ordered chronologically using the start and end dates. Each pair of consecutive job records was treated as one observed occupational transition. For example, if a person moved from occupation A to occupation B, this contributed one transition from A to B. Only transitions for which both the source and destination job records had valid occupational codes were retained. Transition counts were then calculated by summing the number of observed moves from each source occupation to each destination occupation. The same approach was also applied at the broader one-digit occupational-group level for visualization purposes. To describe transition patterns, counts were converted into row-normalized transition shares. This means that, for each source occupation, the number of transitions to each destination occupation was divided by the total number of transitions starting from that source occupation. The resulting value indicates the observed share of workers moving from a given source occupation to each possible destination occupation. These measures are descriptive and reflect observed profile histories rather than population-weighted labour market flows.

## Digital Semantic Score

The digital semantic score is a measure created by the researchers of this paper of how close a job title or skill is to digital concepts, compared with a non-digital reference set. It is based on relative semantic proximity in the embedding space, not on keyword matching and not on a supervised classifier.

In order to quantify these concepts, each job title and skill text, were converted to embeddings using the embeddings model "text-embedding-3-small". First the text was stripped, lowercased and white spaced-normalized, and then the embedding vector was normalized to unit length.

All semantic similarities were computed as cosine similarities, which are dot products between unit-normalized vectors.

First, we defined four small sets of anchor phrases. Three sets represent different types of digital work: *core_digital*, *hybrid_digital*, and *business_systems_automation*. A fourth set, *non_digital_baseline*, represents the non-digital work such as manual labour, hospitality service, retail sales, construction trades, food preparation, cleaning services, agricultural work, physical logistics, and craft occupations.

Each anchor phrase was embedded with the same embedding model as the job titles and skills. All embedding vectors were then normalized to unit length. For each anchor group, we calculated a centroid vector by averaging the unit-normalized vectors of the phrases in that group and normalizing the resulting centroid. For every embedded title or skill, we then calculated cosine similarity to each centroid:

$$\mu_k = \frac{\frac{1}{|A_k|}\sum_{a\in A_k} \tilde{x}_a}{\left\|\frac{1}{|A_k|}\sum_{a\in A_k} \tilde{x}_a\right\|_2},$$

where $A_k$ is the set of anchor phrases in group $k$. The similarity between text $t$ and anchor group $k$ is:

$$s_k(t) = \tilde{x}_t^{\top} \mu_k.$$

The positive digital component is the maximum similarity to the digital anchor groups:

$$s_D(t) = \max_{k\in\mathcal{K}_D}(s_k(t))$$

where $\mathcal{K}_D$ contains the core digital, hybrid digital, and business systems or automation anchor groups. To reduce the risk that generally abstract or professional language is mistaken for digital content, the score is adjusted by subtracting the similarity to the non-digital baseline centroid:

$$D(t) = s_D(t) - s_B(t),$$

where $s_B(t)$ is the similarity to the non-digital baseline. Higher values indicate that a title or skill is semantically closer to the digital anchor space than to the non-digital baseline. The score is interpreted as a relative semantic index, not as a percentage or probability.

This subtraction is important because a title can be broadly similar to many work-related concepts simply because it is a job title. By subtracting the non-digital baseline similarity, the score emphasizes texts that are closer to digital anchors than to general non-digital work anchors.

This means that a title or skill can be classified by its closest digital dimension while still receiving a low or negative final digital score if it is even closer to the non-digital baseline. (For example, a retail title may have its highest digital similarity to *hybrid_digital*, because retail can be semantically related to e-commerce or customer relationship management (CRM), but if it is more similar to *retail sales* and other non-digital baseline anchors, its final digital semantic score will be low or negative.)

The anchor sets are deliberately concise and transparent. They should be interpreted as reproducible semantic reference points rather than a complete taxonomy of all digital and non-digital work. In robustness checks, future versions of the analysis could vary the anchor phrases or add domain-specific anchors, but the present results are based exactly on the four anchor groups listed above.

## Anchor Group Definitions and Rationale

The anchor groups were defined manually as transparent semantic reference sets before calculating the digital semantic scores. They are not estimated from the outcome data and they are not supervised labels. Their purpose is to define interpretable directions in embedding space: one direction for core technical digital work, one for hybrid digital work, one for business systems and automation, and one non-digital baseline direction. All anchor phrases were embedded with the same model as job titles and skills.

The anchor phrases were written in English because the main job-title text used for embedding was translated to English, and the skill strings are normalized into English-like lower-case terms where available. Each anchor phrase is broad enough to represent a work domain rather than a single software tool or narrow occupation. This reduces dependence on one keyword and makes the centroid represent a semantic field.

### *Core Digital*

The *core_digital* anchor group is intended to represent technical digital production and infrastructure work. It captures occupations and skills directly involved in developing, maintaining, securing, or analyzing digital systems and data.

The exact anchor phrases are:

*software engineering, data analytics, artificial intelligence, machine learning, cloud computing, cybersecurity, database engineering, web development, information technology, business intelligence*

These phrases were chosen to cover the main technical digital domains: software development, data and AI, cloud infrastructure, cybersecurity, databases, web systems, IT, and business intelligence. A title or skill close to this centroid is interpreted as semantically close to core digital work.

### *Hybrid Digital*

The hybrid_digital anchor group is intended to represent roles where digital tools, platforms, analytics, or product practices are central, but where the occupation is not necessarily a core technical IT role. This group captures digitally mediated business, marketing, product, and customer-facing work.

The exact anchor phrases are:

*digital transformation, digital marketing, online platforms, product owner, scrum master, agile project management, customer analytics, crm systems, user experience design, e-commerce operations*

These phrases were chosen to cover digital transformation, online business models, agile/product roles, customer analytics, CRM, UX, and e-commerce. A title or skill close to this centroid is interpreted as semantically digital in a hybrid or digitally enabled business sense.

### *Business Systems and Automation*

The business_systems_automation anchor group is intended to represent enterprise software, administrative digitization, process automation, and operational systems. This group captures digitalization through organizational systems rather than through software engineering alone.

The exact anchor phrases are:

*enterprise resource planning, business, process automation, workflow automation, robotic process automation, sap systems, Microsoft dynamics, low-code automation, data-driven operations*

These phrases were chosen to capture ERP, workflow systems, RPA, large enterprise platforms, low-code/no-code automation, and data-driven operations. A title or skill close to this centroid is interpreted as semantically close to business process digitalization or automation.

### *Non-Digital Baseline*

The non_digital_baseline anchor group is intended to represent work domains that are generally less digital in their occupational language. This group is not meant to imply that these jobs never use digital tools. Instead, it provides a baseline semantic direction for traditional manual, service, craft, logistics, and administrative work.

The exact anchor phrases are:

*manual labour, hospitality service, retail sales, construction trades, food preparation, cleaning services, agricultural work, traditional administration, physical logistics, craft occupations*

These phrases were chosen to span broad non-digital or less digitally specific work domains. The baseline is used as a contrast term in the final score. A title can have some similarity to digital anchors simply because all job titles share general work-related language; subtracting similarity to this baseline helps isolate whether the title is more digital than general non-digital work.

The three digital groups are kept separate because digitalization is not one single phenomenon. A core IT occupation, a digital marketing role, and an ERP automation role can all be digital, but they are digital in different ways. The method therefore first measures similarity to each digital domain separately and then uses the strongest of the three digital similarities.

## Job-title digitalization over time

Each job record inherits the digital semantic score of its normalized title. For year $y$ and ISCO major group $m$, the mean job-title digitalization score is:

$$\overline{D}_{my} = \frac{1}{n_{my}} \sum_{i:\text{year}(s_i)=y,\, m_i=m} D(t_i)$$

where $n_{my}$ is the number of observed job records in that year and major group. These averages describe the semantic content of observed titles in the dataset and are not adjusted with external employment weights.

## Four-digit occupational digital drift

To summarize longer-run occupational change, four-digit ISCO occupations are compared between a baseline period and a recent period. The baseline period is 2000-2010 and the recent period is 2020-2024. For occupation $g$ and period $q$, the row-weighted mean title score is:

$$\overline{D}_{gq} = \frac{1}{n_{gq}} \sum_{i:g_i=g, i\in q} D(t_i)$$

where $n_{gq}$ is the number of observed job records in occupation $g$ during period $q$.

The digital drift of occupation $g$ is:

$$\Delta_g = \overline{D}_{g,\text{recent}} - \overline{D}_{g,\text{baseline}}$$

Occupations with low support in either period are flagged or excluded from reportable rankings. Support filtering is used to improve descriptive stability; it should not be interpreted as a formal statistical significance test.

## Transition-level semantic change

For each same-person transition from source job $u$ to destination job $v$, the change in title digitalization is:

$$\delta_{uv} = D(t_v) - D(t_u)$$

Transition summaries aggregate these changes by source and destination occupational groups. For a transition group $A \to B$, the mean semantic shift is:

$$\bar{\delta}_{AB} = \frac{1}{n_{AB}} \sum_{(u,v):m_u=A,\, m_v=B} \delta_{uv}$$

Balanced transition visualizations report positive, near-zero, and negative transition groups to avoid presenting only the largest increases.

## Skill-level analysis

Skills are treated as person-level attributes rather than job-specific attributes. Each person's skill list is parsed into distinct normalized skill strings. Individual skills receive the same anchor-based Digital Semantic Score $D(t)$ as job titles. A person-level average skill digitalization score can be written as:

$$H_p = \frac{1}{|S_p|} \sum_{s \in S_p} D(s),$$

where $S_p$ is the set of parsed skills for person $p$. Because skill lists are not associated with each specific job, skill analyses are interpreted as profile-level associations rather than direct measures of year-specific occupational skill requirements.

## Results & Discussion

The following section presents several results from the methods described above.

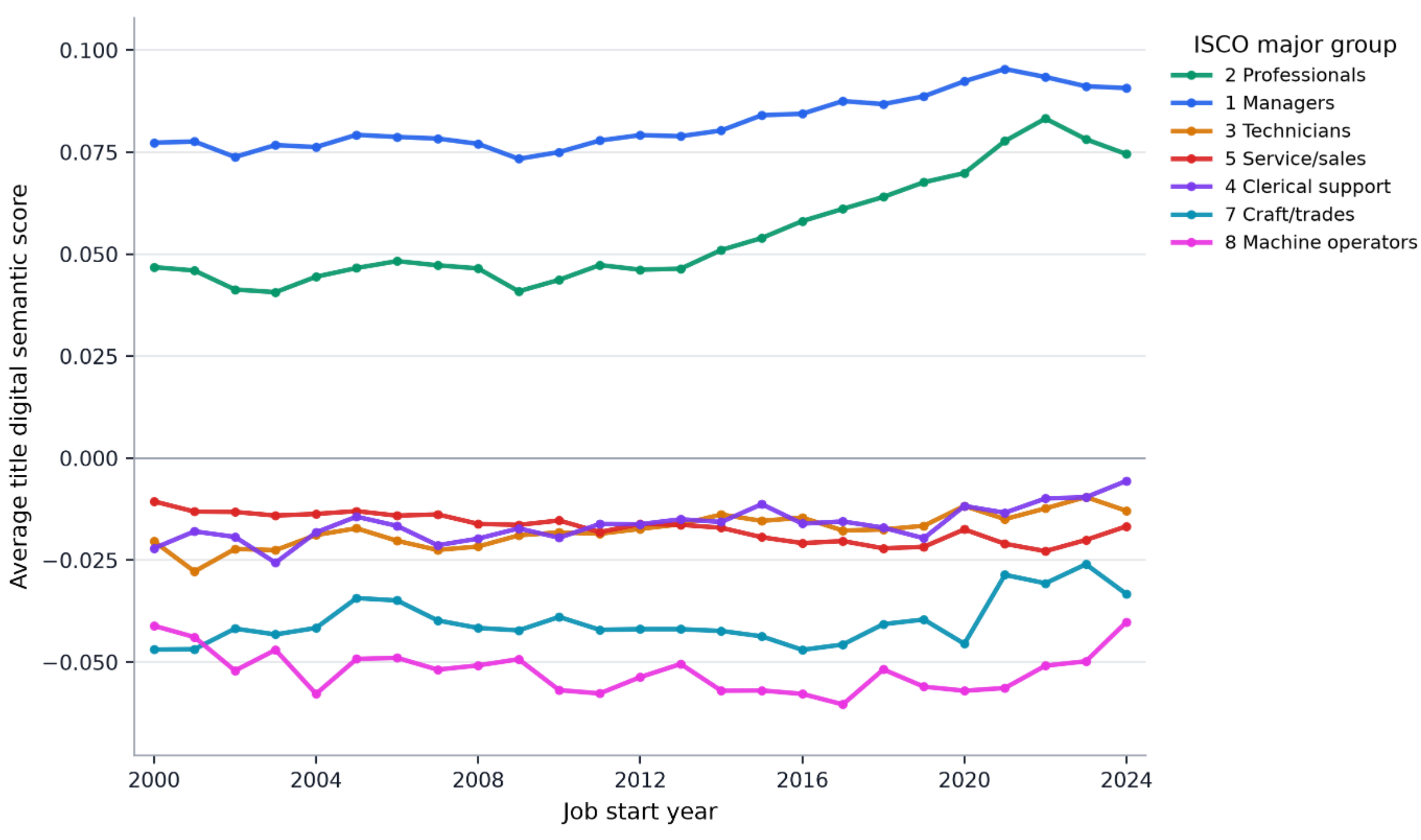

Annual means are computed at job-row level; years with fewer than 100 scored rows in a major group are omitted.

**Figure 1.** Long-Run Digitalization by ISCO Major Group

Figure 1 plots annual average digital semantic scores of job-title language by ISCO major group. The x-axis represents job start year, while the y-axis reports the mean digital semantic score across all scored jobs in each year-major-group. Colors distinguish ISCO major groups. The figure shows whether broad occupational groups have become more digitally oriented over time, with upward trajectories indicating increasing digital semantic intensity in observed job-title language.

Figure 1 shows that the digital semantic content of job-title language changed over time for some of the major groups. Managers and professionals consistently have the highest average digital scores, and both groups show a gradual rise after the mid-2010s, suggesting that digital terminology has become more embedded in higher-skill and knowledge-intensive occupations. This finding is consistent with previous research showing that the effects of digitalization and AI are unevenly distributed across occupations with the more knowledge-based and professional occupations showing higher levels of digital skill demand and technology exposure (Green, 2024; Engberg et al., 2024).

By contrast, technicians, service and sales workers, clerical support workers, craft workers, and machine operators remain close to or below zero for most of the period. This does not mean that these occupations are unaffected by digitalization; rather, their job titles remain semantically closer to the non-digital baseline than to the digital anchor groups. Overall, the figure suggests that digitalization is not a uniform labour market shift. It appears more strongly in the language of managerial and professional work, while other occupational groups show weaker or a more gradual semantic change.

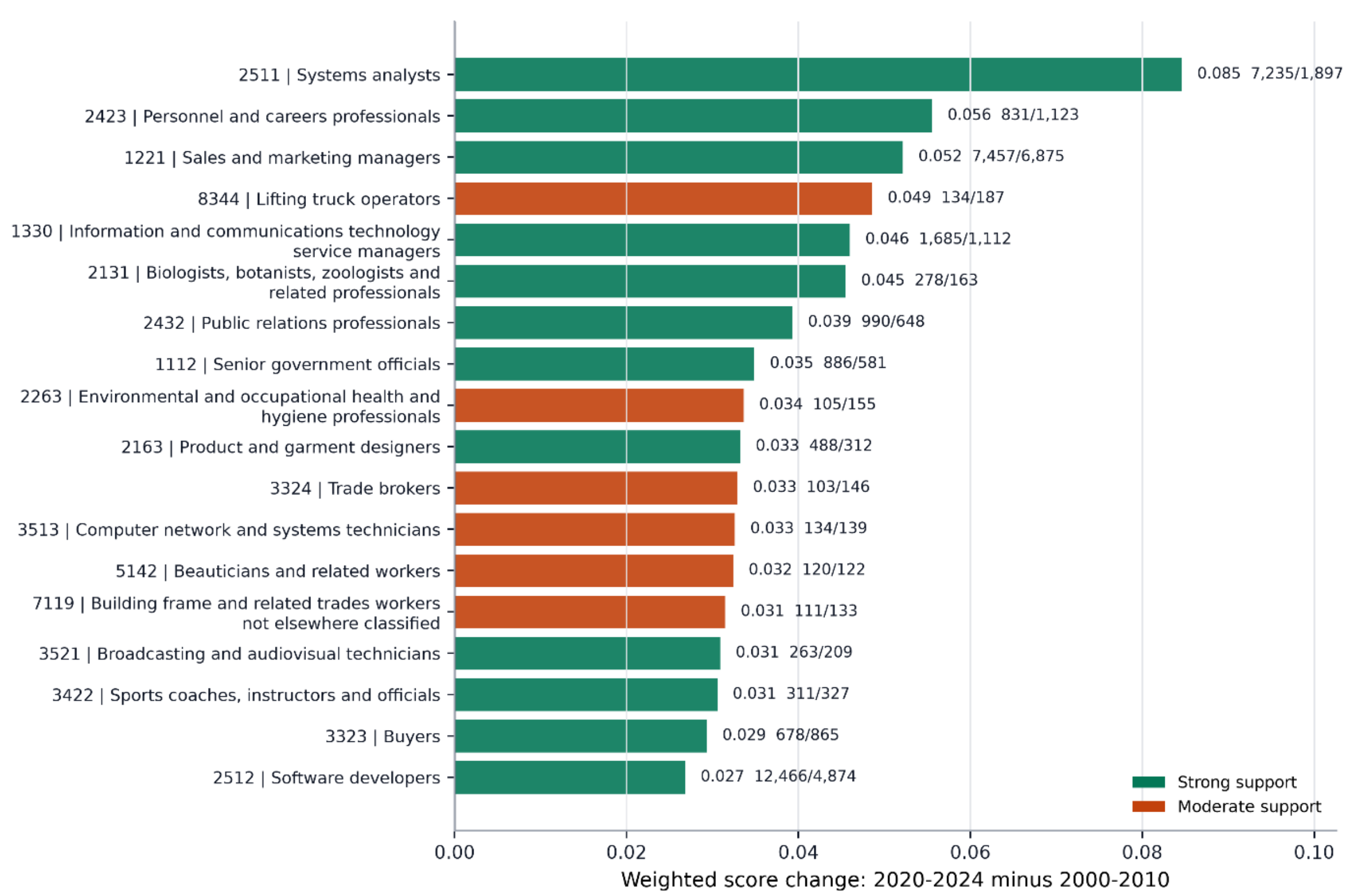

**Figure 2.** Occupations with the Largest Increases in Digital Job-Title Language

Figure 2 ranks 4-digit ISCO occupations by their reportable increase in job-title digital semantic score between 2000-2010 and 2020-2024. The x-axis gives the job-row-weighted score change, and the y-axis lists the corresponding ISCO 4-digit occupations. Bar length represents the magnitude of the increase. Color distinguishes support categories, with green denoting strong support (if recent job rows were at least 250 and baseline job rows were at least 100) and orange denoting moderate support (if recent job rows were at least 100 and baseline job rows were at least 30). Labels report both the score change and the recent/baseline job-row counts, allowing the magnitude of change to be interpreted together with empirical support.

The strongest increases are concentrated in professional and managerial occupations, especially ICT-related and technical-professional groups such as ISCO 2511, 2423, 1221, and 2131. This suggests that digitalization is not only visible in explicitly technical jobs, but also in management, consulting, administration, and specialized professional roles where digital tools, data systems, and platform-based work have become more central to occupational language. This finding is aligned with the concept of skill rebundling, where the digital technologies reshape existing occupations through the new sequences of technical, analytical and domain-specific skills rather than through the emergence of totally new occupations (Stephany, 2021).

The figure also shows that some non-professional groups display meaningful increases, including machine operators, technicians, service/sales, and craft/trade occupations. These cases are important because they suggest that digital language is spreading beyond the highest-scoring occupational groups, although often with lower support or smaller changes. The largest changes with strong support are more reliable for interpretation, while moderate-support rows should be treated as suggestive. Overall, Figure 2 supports the view that digitalization is occupationally uneven, with the clearest robust gains concentrated in knowledge-intensive and managerial work, but with signs of diffusion into selected technical and operational occupations.

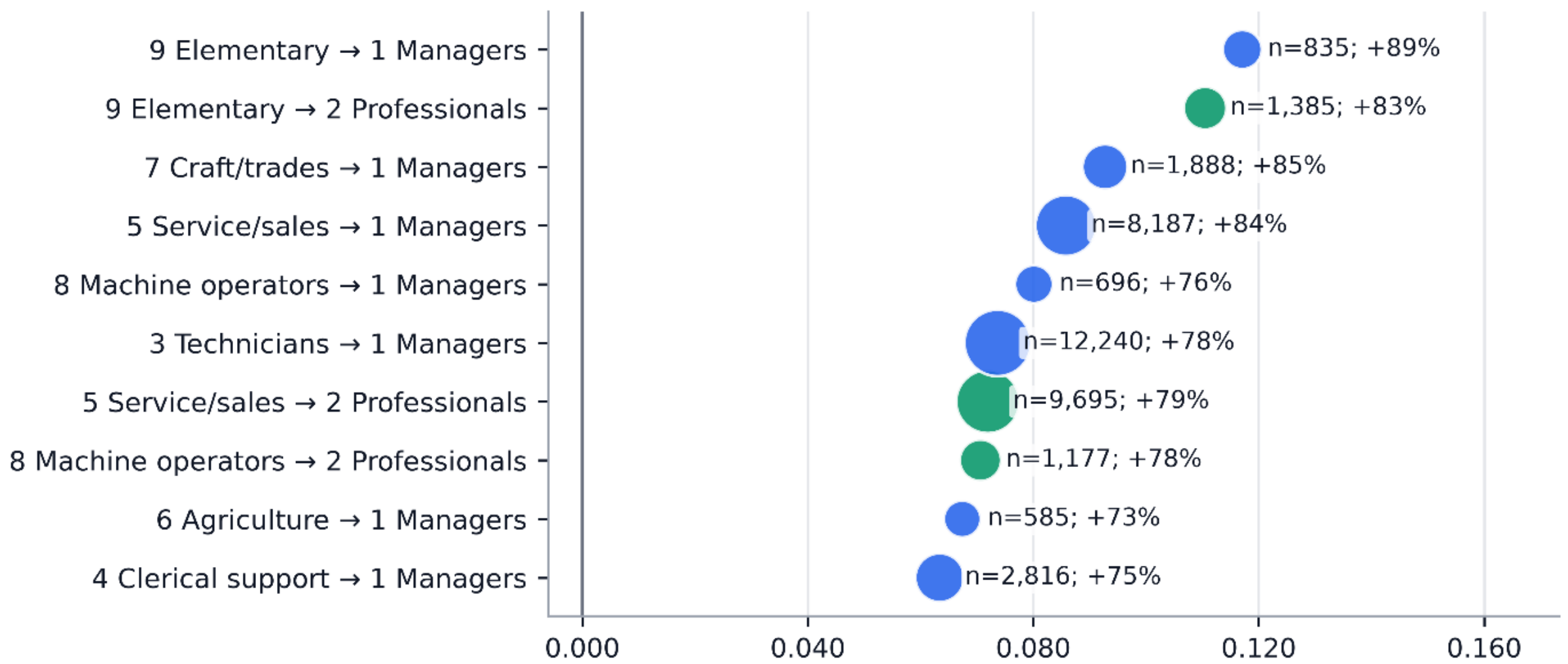

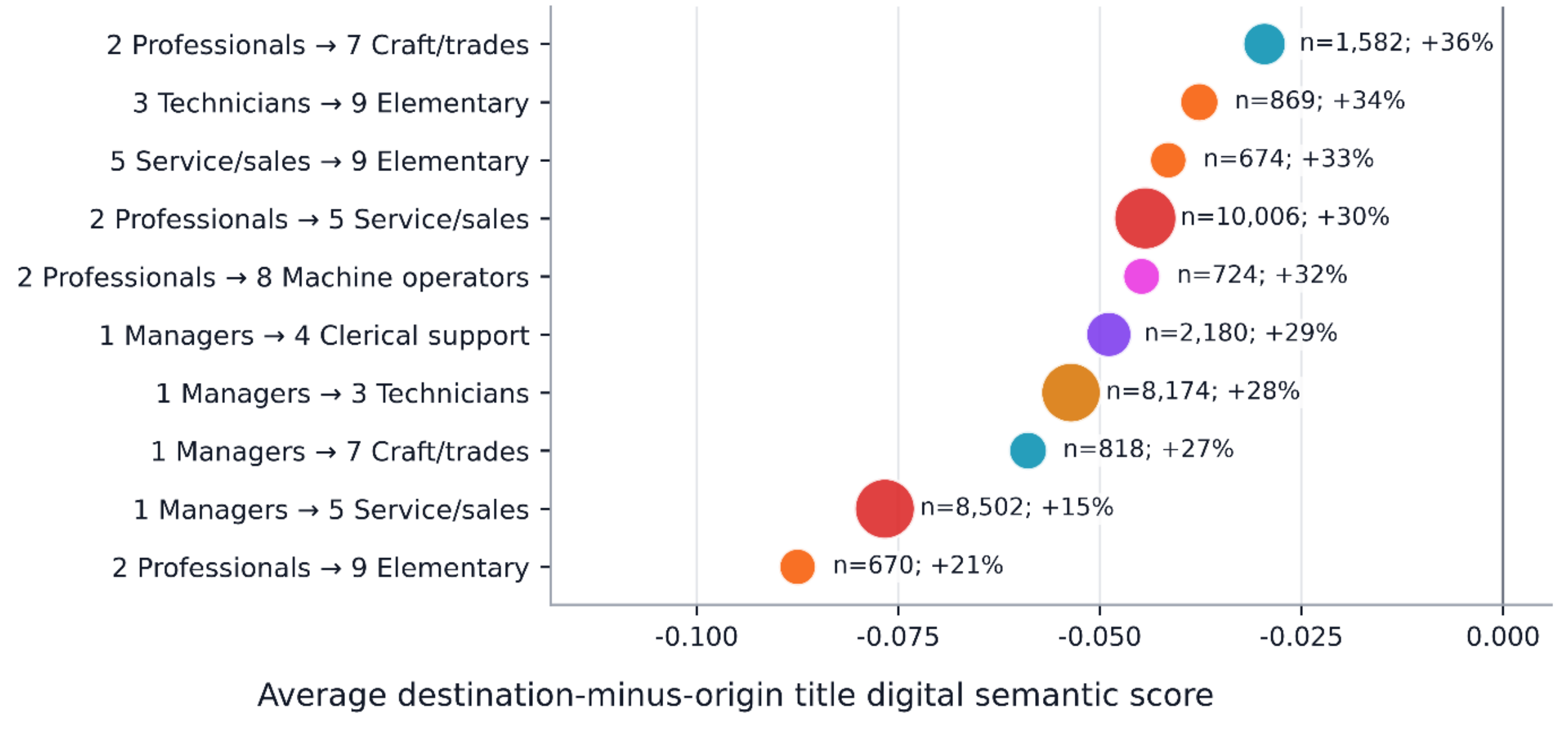

**Figure 3.** Transition Pathways and Digital Score Shifts

Figure 3 examines whether observed career transitions are associated with movement toward more or less digital job-title language. Each point represents a transition pathway between an origin ISCO major group and a destination ISCO major group. The x-axis reports the average change in digital semantic score between the destination and origin job title, calculated as the destination title score minus the origin title score. Positive values therefore indicate transitions into more digitally oriented job-title language, while negative values indicate transitions into less digitally oriented job-title language. Point size represents the number of scored transitions in the pathway, and colour identifies the destination ISCO major group. The labels report both the transition count and the share of individual transitions with a positive score change.

The figure separates the transition pathways into two panels: the ten largest positive shifts, and the ten lowest or negative shifts. This design shows that digitalization through mobility is not uniform. Some career pathways are strongly associated with movement toward more digital titles, while others show essentially no average change or even movement toward less digital title language.

The largest positive shifts are concentrated in pathways into managers and professionals. For example, transitions from elementary occupations to managers show the strongest average increase in title digital score, with an average shift of approximately 0.117 and 89% of individual transitions showing a positive score change. Transitions from elementary occupations to professionals also show a large positive shift, approximately 0.111, with 83% positive transitions. Other large positive pathways include craft and related trades workers to managers, service and sales workers to managers, technicians to managers, and service and sales workers to professionals. These patterns suggest that upward or lateral moves into managerial and professional destinations are often accompanied by more digitally oriented occupational language.

The lowest or negative shifts show the opposite pattern. Transitions from professionals to elementary occupations have the strongest negative average change, approximately -0.088, with only 21% of transitions showing a positive score change. Transitions from managers to service and sales workers also show a large negative shift, approximately -0.077, with only 15% positive transitions. Other negative pathways include managers to craft and related trades, managers to technicians, managers to clerical support, professionals to machine operators, and professionals to service and sales workers. These patterns suggest that moves away from managerial or professional destinations toward service, elementary, craft, clerical, or machine-operating destinations are often associated with less digitally oriented job-title language.

Overall, Figure 3 shows that digitalization is not only an occupation-level trend but is also visible in mobility pathways. Some transitions move workers into more digital occupational language, especially when the destination is managerial or professional. Other pathways are neutral or negative. The result therefore supports a heterogeneous view of labour market digitalization: career mobility can be associated with both upward digital semantic shifts and downward or non-digital shifts, depending on the origin and destination occupational groups. These patterns should be interpreted descriptively, as associations in observed job histories, rather than as causal effects of mobility on digital skill acquisition or job content.

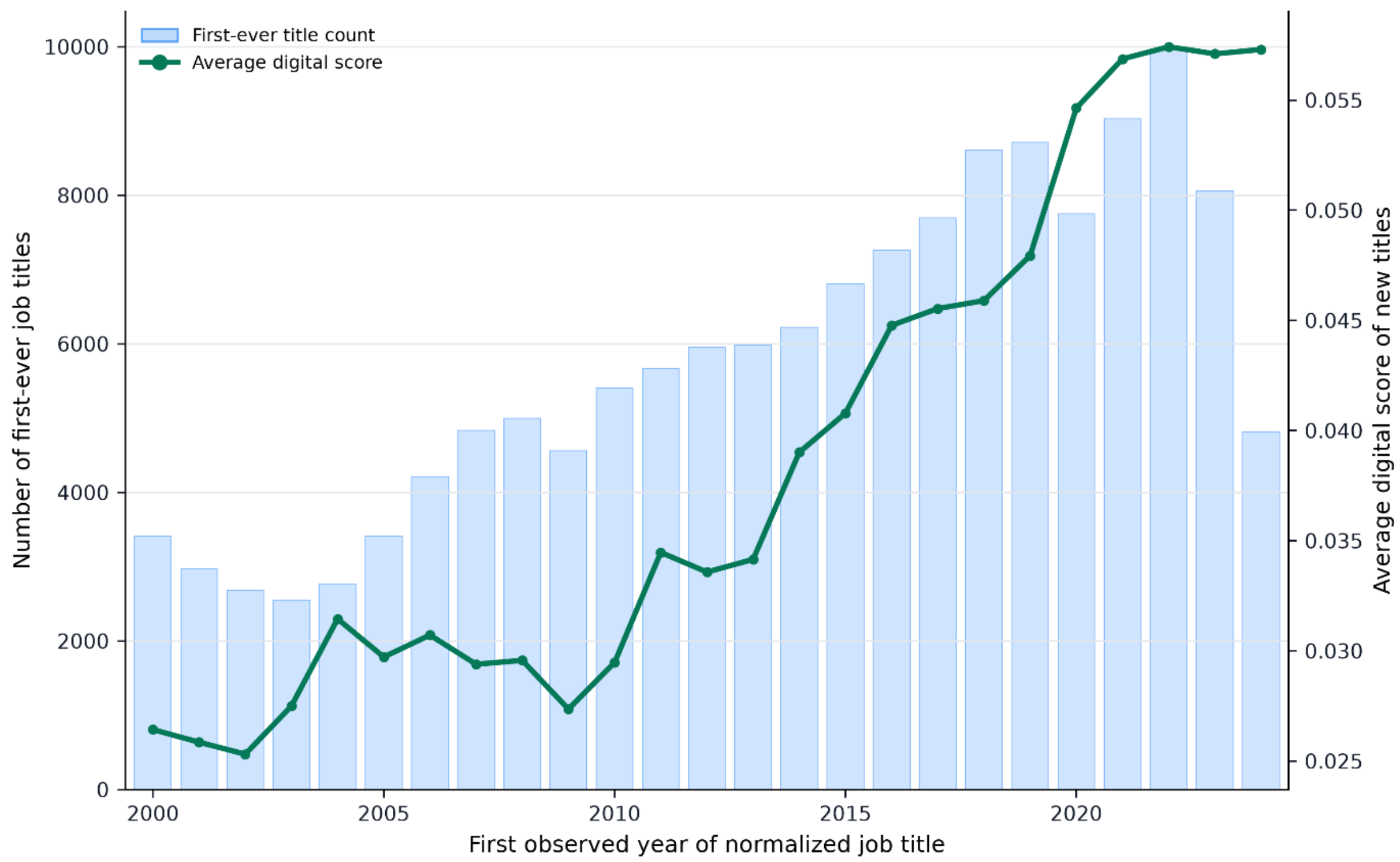

**Figure 4.** Digital Semantics Of Newly Emerging Job Titles

Figure 4 examines the digital semantic content of newly emerging job-title vocabulary. To separate vocabulary emergence from job-row volume, each normalized job title is counted only once, in the year it first appears in the dataset. The x-axis shows the first observed year of the title. The blue bars, read against the left y-axis, show the number of first-ever titles appearing in each year. The green line, read against the right y-axis, shows the average digital semantic score of those newly observed titles.

The main pattern is that newly appearing job titles become increasingly digital over time. From the early 2010s onward, the average digital semantic score of new titles rises clearly and remains high in the most recent years. This suggests that the frontier of occupational vocabulary is increasingly shaped by digital concepts, tools, platforms, analytics, automation, and digitally mediated work practices. This finding is in support of previous research that nods to the rapid emergence of new digital skills and occupational terminology that may not be yet fully captured by existing occupational and skills taxonomies (Napierała, 2024).

This figure complements the occupation-level drift shown in Figure 2. While Figure 2 asks whether language within established four-digit occupations has become more digital over time, Figure 4 focuses on the entry of new occupational labels. The rising digital score of new titles indicates that digitalization is not only visible through changes within existing occupations, but also through the creation and diffusion of new job-title language. In other words, digitalization appears as a process of occupational language expansion: new roles, specializations, and labels increasingly encode digital capabilities.

This ability to identify digitalisation through emerging occupational language also complements the taxonomy-based approaches shown by Lennon et al. (2023), which rely on predefined digital skills contained within ESCO. In comparison, since the Digital Semantic Score is based on semantic similarity rather than predefined digital skill lists, it can detect digitalisation signals in both new job titles and hybrid occupational terminology even when these are not yet explicitly classified as digital skills.

The joint movement of the bars and the line is also informative. Over much of the period, the dataset contains a growing variety of newly observed job titles, and these new titles are increasingly digital in their semantic content. This points to both occupational differentiation and digital intensification. The labour market is not only generating more distinct labels; the semantic direction of those labels increasingly points toward digital work. Importantly, this pattern is not limited to core ICT occupations. It also fits the broader evidence that digital terminology is spreading into business, management, marketing, administration, and service-oriented professional work.

The later years should be interpreted with care. The decline in the number of first-ever titles in 2024 likely reflects partial-year coverage, data sparsity, and incomplete observation rather than a real decline in occupational innovation. However, the average digital score remains high even in these later years.

A methodological strength of this approach is that repeated occurrences of common job titles do not dominate the measure. By counting each normalized title only once, Figure 4 captures the semantic score of newly entering title types rather than the volume of job rows. At the same time, this is also a limitation. The figure does not show how many people held each new title, how long the title persisted, or whether it became common in the labour market. A rare title and a widely adopted title both enter the title-count measure once.

Overall, Figure 4 adds evidence that digitalization is visible at the frontier of occupational naming. Together with the occupation-level drift and transition analyses, it supports the interpretation that digitalization is not a single shift concentrated only in ICT jobs, but a broader semantic transformation in how work is described, differentiated, and organized.

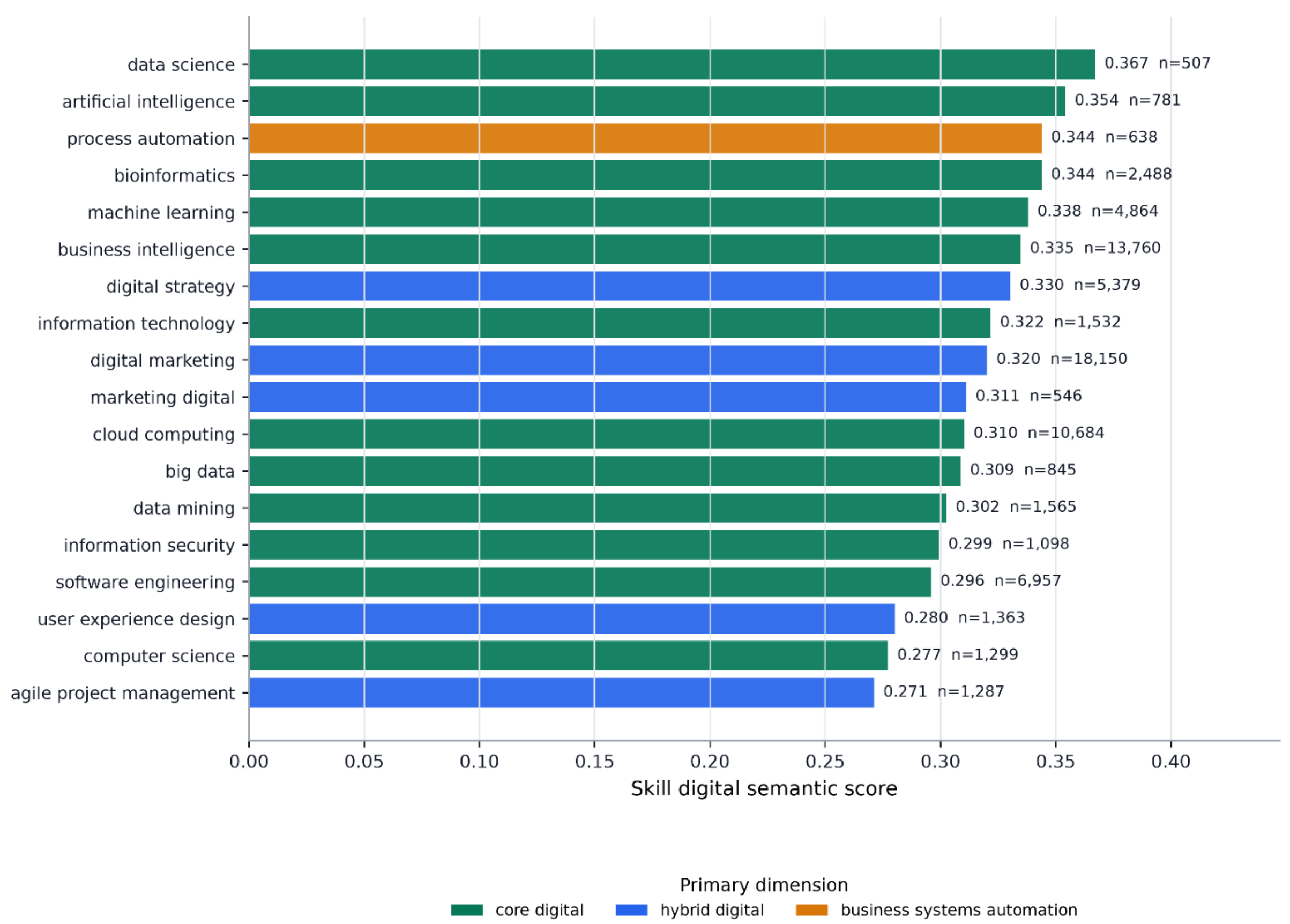

**Figure 5**: High-Prevalence Digital Skills by Semantic Score

Figure 5 ranks high-support person-level skills by their digital semantic score. The y-axis lists individual skills, and the x-axis reports each skill's semantic proximity to digital concepts relative to the non-digital baseline. Bar length represents the magnitude of the skill's digital semantic score. Colours indicate the closest digital anchor dimension: core digital, hybrid digital, or business systems and automation. Labels report both the score and the number of people listing the skill. Because skills are observed at person level, the figure should be interpreted as a profile-level view of digital skill content rather than evidence of skill use in a specific job or year.

The figure ranks high-prevalence skills by their digital semantic score, keeping only skills listed by at least 500 people. This support threshold is important because it focuses on interpretation skills that are not only semantically digital but also sufficiently common in the observed profiles. The figure therefore highlights broad digital capability signals rather than rare or highly specialized skill terms.

The strongest-scoring skills are dominated by core digital capabilities. Skills such as data science, artificial intelligence, bioinformatics, machine learning, business intelligence, information technology, cloud computing, big data, data mining, information security, software engineering, and computer science all appear near the top. This pattern confirms that the embedding-based score behaves as expected: skills directly associated with software, data, AI, security, cloud systems, and computational work receive the highest digital semantic values. This corresponds with previous studies showing persistent demand for AI, data, software and cloud-related skills across digital occupations (Beblavý et al., 2016; Green, 2024).

At the same time, the figure shows that digital capability is not limited to technical programming or IT infrastructure. Some of the high-ranking skills fall into hybrid digital and business-systems domains, including digital strategy, digital marketing, marketing digital, user experience design, agile project management, and process automation. These skills point to the diffusion of digital work into business, design, marketing, management, and organizational transformation. In other words, digitalization appears not only as technical production but also as digitally enabled coordination, customer engagement, workflow redesign, and platform-based business practice.

The distinction between the three primary dimensions is analytically useful. The core digital group captures technical and analytical capabilities such as data science, AI, machine learning, cloud computing, and software engineering. The hybrid digital group captures skills that connect digital tools with organizational and market-facing work, such as digital strategy, digital marketing, user experience design, and agile project management. The business systems automation group is represented by process automation, pointing to enterprise and operational digitalization. This supports the broader claim that digitalization is multidimensional rather than a single homogeneous skill category.

The prevalence counts also add an important layer of interpretation. Some highly digital skills are relatively specialized, such as data science or artificial intelligence, while others are much more widely reported, such as digital marketing, cloud computing, business intelligence, software engineering, and machine learning. This means that a skill's semantic digital intensity and its diffusion across profiles are distinct. A skill can be highly digital but relatively narrow in prevalence, or somewhat lower in score but more broadly distributed. Figure 5 shows both dimensions together: the semantic score and the number of people reporting the skill.

The presence of digital marketing and digital strategy among the high-prevalence digital skills is especially important for the interpretation of the labour market transition figures. It helps explain why some non-ICT occupations and hybrid professional roles can have high digital semantic scores. Workers in advertising, marketing, management, consulting, product, and administrative roles may not report core programming skills, but they may report digital strategy, CRM, analytics, platforms, user experience, digital marketing, or process automation skills. These skills can raise the digital profile of occupations that are not classified as ICT occupations in a narrow sense.

Figure 5 also helps interpret why average skill scores can sometimes look similar across different occupations. Two occupations may have similar average digital skill scores while relying on different types of digital capability. For example, one group may be characterized by software engineering, AI, and cloud computing, while another may be characterized by digital marketing, user experience, agile project management, and digital strategy. The scalar digital score captures overall digital intensity, but the primary-dimension categories reveal that the underlying content can differ.

A key caveat is that the skills in this analysis are person-level profile attributes, not job-specific observations. The figure should therefore not be interpreted as showing which skills were used in a particular job, acquired in a specific year, or required by a specific employer. Instead, it shows which reported skills are semantically closest to digital anchor concepts among skills that are common enough to support stable interpretation. The figure is strongest as evidence about the digital content of worker skill profiles, not as a direct measure of job-specific skill demand.

Overall, Figure 5 supports the main argument that digitalization is visible across several layers of labour market information. Job titles show changing occupational language; transition figures show movement between more and less digital occupational contexts; and the skill figure reveals the capability domains underlying those patterns. The dominance of core digital skills confirms the technical foundation of digitalization, while the presence of hybrid and automation-related skills shows how digital capability extends into broader professional, managerial, marketing, design, and organizational work.

## Conclusion

This article shows that digitalisation in the Dutch labour market is not a uniform process, but one that differs across occupations, career pathways and skill profiles. By combining large-scale online profile data with AI-supported occupational classification and semantic scoring, the analysis provides a more detailed view of how digital work is emerging, reshaping and spreading. The results indicate that digitalisation is most visible in managerial, professional and ICT-related occupations, but also increasingly appears in hybrid business, marketing, administrative and automation-related roles. The findings further show that digitalisation is reflected in the evolution of occupational language, with newly emerging job titles displaying increasingly digital characteristics.
Career transitions further show that movement into more digital work is pathway-dependent, while the skill-level analysis highlights the multidimensional nature of digital capability.

A significant contribution of this article is the introduction of the Digital Semantic Score, which enables the measurement of the spread of digitalisation through occupational language and worker skill profiles rather than having to rely on the predefined occupational groups or keyword-based indicators.

The findings further demonstrate that digital transformation is visible not only within occupations, but also in the emergence of new occupational vocabulary and in mobility pathways connecting occupations with different digital profiles. Together, these results suggest that digitalisation should be understood as a broad labour market transformation extending beyond traditional ICT occupations and into hybrid professional, managerial and operational roles.

Several limitations of the data and methodology should be considered when interpreting the results.
First, the underlying dataset is derived primarily from LinkedIn user profiles and therefore does not constitute a representative sample of the Dutch workforce. LinkedIn users tend to be overrepresented among highly educated, white-collar, professional, managerial, and knowledge-intensive occupations, while workers in manual, routine, lower-skilled, or less digitally oriented occupations are generally underrepresented. Although the large scale of the dataset enables detailed analyses of occupational language, skills, and career transitions, the results may not fully capture digitalisation dynamics among all segments of the workforce.

Second, an important limitation concerns the structure of the skill data. Skills are observed only at the individual profile level and cannot be linked to specific jobs within a person's employment history. At most, it indirectly provides information on which skills the owners of the profiles believe they need to be attractive to employers and clients (the demand site of the labour market). As a result, skills are treated as relatively static attributes of individuals rather than as occupation-specific characteristics that evolve over time. This means that it is not possible to determine which skills were used in a particular job, when a skill was acquired, or how skill portfolios changed across career transitions. Consequently, the skill analyses should be interpreted as reflecting the digital orientation of individuals' overall skill portfolios rather than the changing skill requirements of specific occupations or jobs.

Third, the skill information is self-reported by users. Self-reported skills are subject to several sources of measurement error. Profiles may be incomplete, outdated, or selectively maintained, leading to the omission of relevant skills. In addition, individuals may differ in how they describe similar competencies, resulting in inconsistencies in skill labels. Some users may also overstate or understate their expertise. Although the use of normalized skill strings and semantic embeddings reduces part of this variation, self-reported skills remain an imperfect proxy for actual skill possession and utilization.

While future research could further examine the evolution of skills at the job level, the proposed methodology provides a scalable framework for monitoring digital labour market change, identifying emerging skill needs, and supporting evidence-based reskilling and labour market policies in the Netherlands.

## Acknowledgments

This article is published as part of Deliverable 2.2 ("Dataset online platforms") of the SkiLMeeT project. The SkiLMeeT project is funded by the European Union's Horizon Europe Research and Innovation program under grant agreement No 101132581. Views and opinions expressed are however those of the author(s) only and do not necessarily reflect those of the European Union or European Research Executive Agency. Neither the European Union nor the granting authority can be held responsible for them.

## Supplementary Files

**Appendix 1** - Changes in Occupational Digital Skill Profiles Across 4-Digit Career Transitions

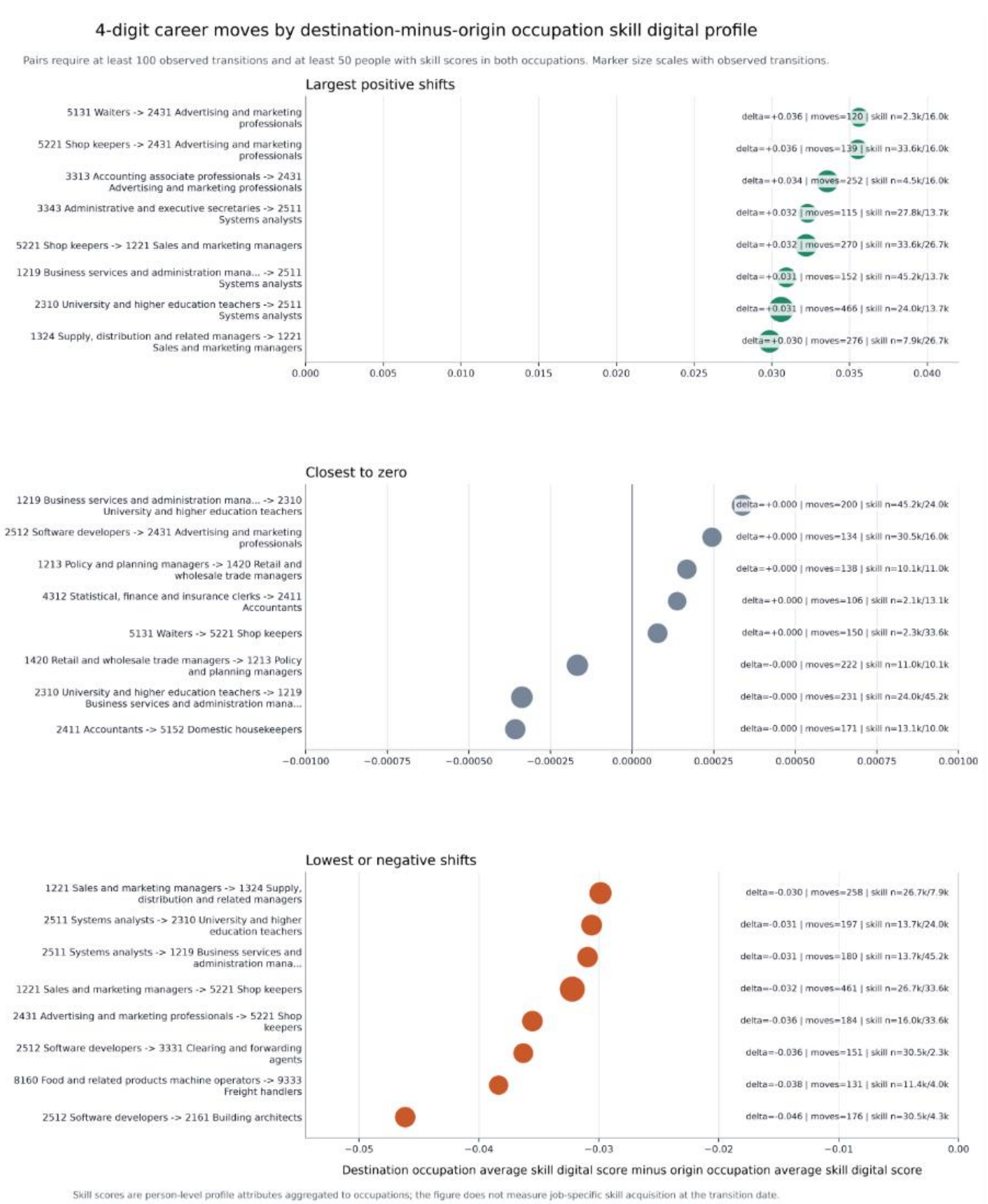

**Figure S1.** Changes in Occupational Digital Skill Profiles Across 4-Digit Career Transitions.

Figure S1 examines whether observed 4-digit ISCO career moves connect occupations with different average digital skill profiles. Unlike Figure 3, which used the digital semantic score of job titles before and after a transition, this figure uses person-level skill embeddings aggregated to occupations. For each 4-digit ISCO occupation, we calculated the average digital semantic score of the skills reported by workers observed in that occupation. Each transition is then summarized as the destination occupation's average skill digital score minus the origin occupation's average skill digital score. Positive values therefore indicate movement toward occupations whose workers have more digitally oriented skill profiles on average.

The figure shows three types of supported 4-digit transition pathways: the largest positive skill-profile shifts, pathways close to zero, and the lowest or negative shifts. All displayed pairs have at least 100 observed transitions and at least 50 workers with skill scores in both the origin and destination occupations.

The positive shifts suggest several plausible routes into more digitally oriented occupational environments. Moves into advertising and marketing professionals, sales and marketing managers, and systems analysts stand out. These destinations are not all "core IT" occupations in a narrow sense. Advertising, marketing, sales management, and business systems roles often combine domain expertise with digital tools, analytics, platforms, CRM systems, automation, or data-driven decision-making. This supports the broader interpretation that digitalization is not confined to software and ICT occupations. It also appears in hybrid professional and managerial roles where digital capabilities are embedded in commercial, administrative, and organizational work. These patterns are coherent with research suggesting that digital transformation is correlated with changing skill requirements and increasing demand for transferable digital competencies across occupational boundaries (Green, 2024).

The figure also shows that some transitions into more digital skill environments originate in occupations that may not traditionally be considered digital, such as waiters, shop keepers, accounting associate professionals, and administrative secretaries. This suggests that digital mobility pathways may not only involve movement from one technical occupation to another. They may also include transitions from service, retail, clerical, or administrative work into occupations where digital platforms, analytics, systems, or digital communication are more central. In policy terms, this suggests that digital upskilling opportunities may be relevant beyond the obvious STEM or ICT pipeline. This interpretation is also consistent with emerging applied labour market tooling, such as LinkedIn's Career Explorer[1], which uses skill similarity between occupations in order to identify potential career pathways and reskilling opportunities. While the current study is different in that it does not showcase individual career recommendations, it similarly shows that occupational mobility can be analyzed through the relationship between workers skill profiles and destination occupations.

[1] https://linkedin.github.io/career-explorer/

The near-zero transitions show that occupational mobility does not automatically imply movement along a digital skill gradient. Some career moves connect occupations with similar average skill digitality, even when the occupations differ in status, sector, or task content. For example, transitions between software developers and advertising and marketing professionals are close to zero in this metric because both occupations have relatively high average digital skill profiles. Similarly, movements between policy/planning, retail management, accounting, and administrative occupations can occur within a relatively similar skill-digitality band. This helps avoid an overly simple interpretation in which every career transition is treated as either "toward" or "away from" digitalization. The near-zero shift should not be read as evidence of identical skill profiles. Rather, both occupations (let's say software developers and marketing professionals) appear to occupy a similarly high range of average skill digitality, although the underlying composition likely differs: software development is closer to core technical digital work, whereas advertising and marketing professionals may draw more on hybrid digital, platform, analytics, CRM, and digital communication skills. This illustrates a limitation of scalar average scores: they summarize intensity but can hide differences in the type of digital capability.

The negative shifts highlight that mobility is multidirectional. Workers also move from occupations with high average digital skill profiles into occupations with lower average digital skill profiles. Some of these cases may reflect career changes, management or domain shifts, coding noise, temporary work, or the fact that a person's skill profile remains broad even when their destination occupation is less digitally characterized. These negative pathways are important because they show that digitalization is not a one-way ladder. Labour market movement can redistribute digitally skilled workers across less digital occupations as well as draw workers into more digital ones.

The main methodological caveat is that the skills are person-level attributes, not associated for each job. Therefore, Figure S1 should not be interpreted as showing that workers gained or lost digital skills at the time of transition. The figure instead describes the average skill digitality associated with the origin and destination occupations. In other words, it shows whether observed career moves connect occupations embedded in more or less digitally skilled worker profiles. This makes the figure strongest as evidence of occupational skill-context differences, not as evidence of causal skill acquisition.

Overall, Figure S1 suggests that digital labour market change is structured through specific occupational pathways rather than a uniform shift affecting all mobility in the same way.